\documentclass{article}
\usepackage[accepted]{icml2009}
\usepackage{mlapa}
\usepackage{amsmath}
\usepackage{amssymb}
\usepackage{amsthm}
\usepackage{amsfonts}
\usepackage{graphicx}
\usepackage{floatflt}
\usepackage{graphicx}
\usepackage{haldefs}

\newcommand{\searn}{\textsc{Searn}}

\begin{document}
\twocolumn[
\icmltitle{Unsupervised Search-based Structured Prediction}
\icmlauthor{Hal Daum\'e III}{me@hal3.name}
\icmladdress{School of Computing, University of Utah,
             Salt Lake City, UT 84112}
\vskip 0.3in
]

\begin{abstract}
  We describe an adaptation and application of a search-based
  structured prediction algorithm ``\searn'' to unsupervised learning
  problems.  We show that it is possible to reduce unsupervised
  learning to supervised learning and demonstrate a high-quality
  unsupervised shift-reduce parsing model.  We additionally show a
  close connection between unsupervised \searn\ and expectation
  maximization.  Finally, we demonstrate the efficacy of a
  semi-supervised extension.  The key idea that enables this
  is an application of the \emph{predict-self} idea for
  unsupervised learning.
\end{abstract}

\section{Introduction}

A prevalent and useful version of unsupervised learning arises when
both the observed data and the latent variables are structured.
Examples range from hidden alignment variables in speech recognition
\cite{rabiner89tutorial} and machine translation
\cite{brownetal93,vogel96}, to latent trees in unsupervised parsing
\cite{paskin01bigrams,klein04induction,smith05induction,titov07latent},
and to pose estimation in computer vision \cite{ramanan05pose}.  These
techniques are all based on probabilistic models.  Their applicability
hinges on the tractability of (approximately) computing latent
variable expectations, thus enabling the use of EM
\cite{dempster77em}.  In this paper we show that a recently-developed
\emph{search-based} algorithm, \searn\ \cite{daume09searn} (see
Section~\ref{sec:searn}), can be utilized for unsupervised structured
prediction (Section~\ref{sec:unsearn}).  We show: (1) that under an
appropriate construction, \searn\ can imitate the expectation
maximization (Section~\ref{sec:em}); (2) that unsupervised \searn\ can
be used to obtain competitive performance on an unsupervised
dependency parsing task (Section~\ref{sec:gi}); and (3) that
unsupervised \searn\ naturally extends to a semi-supervised setting
(Section~\ref{sec:semi}).  The key insight that enables this work is
that we can consider the prediction of the (observed) input to be,
itself, a structured prediction problem.

\section{Structured Prediction} \label{sec:sp}

The \emph{supervised} structured prediction problem is the task of
mapping inputs $x$ to complex structured outputs $y$ (e.g., sequences,
trees, etc.).  Formally, let $\cX$ be an arbitrary input space and
$\cY$ be structure output space.  $\cY$ is typically assumed to
\emph{decompose} over some smaller substructures (e.g., labels in a
sequence).  $\cY$ comes equipped with a loss function, often assumed
to take the form of a Hamming loss over the substructures.  Features
are defined over pairs $(x,y)$ in such a way that they obey the
substructures (e.g., one might have features over adjacent label pairs
in a sequence).  Under strong assumptions on the structures, the loss
function and the features (essentially ``locality'' assumptions), a
number of learning algorithms can be employed: for example,
conditional random fields \cite{lafferty01crf} or max-margin Markov
networks \cite{taskar05mmmn}.

A key difficulty in structured prediction occurs when the output space
$\cY$, the features, or the loss, does not decompose nicely.  All of
these issues can lead to intractable computations at either training
or prediction time (often both).  An attractive approach for dealing
with this intractability is to employ a search-based algorithm.  The
key idea in search-based structured prediction is to first decompose
the output $y$ into a sequence of (dependent) smaller predictions
$y_1, \dots, y_T$.  These may each be predicted in turn, with later
predictions dependent of previous decisions.

\subsection{Search-based Structured Prediction} \label{sec:sbsp}

A recently proposed algorithm for solving the structured prediction
problem is \searn\ \cite{daume09searn}.  \searn\ operates by
considering each substructure prediction $y_1, \dots, y_T$ as a
classification problem.  A classifier $h$ is trained so that at time
$t$, given a feature vector, it predict the best value for $y_t$.  The
feature vector can be based on any part of the input $x$ and any
previous decision $y_1, \dots, y_{t-1}$.  This introduces a
chicken-and-egg problem.  $h$ should ideally be trained so that it
makes the best decision for $y_t$ \emph{given} that $h$ makes all past
decisions $y_1, \dots, y_{t-1}$ and all future decisions $y_{t+1},
\dots, y_T$.  Of course, at training time we do not have access to $h$
(we are trying to construct it).  The solution is to use an iterative
scheme.

\subsection{\searn} \label{sec:searn}

The presentation we give here differs slightly from the original
presentation of the \searn\ algorithm.  Our motivation for straying
from the original formulation is because our presentation makes more
clear the connection between our unsupervised variant of \searn\ and
more standard unsupervised learning methods (such as standard
algorithms on hidden Markov models).

Let $\cD^{\text{SP}}$ denote a distribution over pairs $(x,y)$ drawn
from $\cX \times \cY$, and let $\ell(y,\hat y)$ be the loss associated
with predicting $\hat y$ when the true answer is $y$.  We assume that
$y \in \cY$ can be decomposed into atomic predictions $y_1, \dots,
y_T$, where each $y_t$ is drawn from a discrete set $Y$.  A
\emph{policy}, $\pi$, is a (possibly stochastic) function that maps
tuples $(x, y_1, \dots, y_{t-1})$ to atomic predictions $y_t$.

The key ingredient in \searn\ is to use the loss function $\ell$ and a
``current'' policy $\pi$ to turn $\cD^{\text{SP}}$ into a distribution
over cost-sensitive (multiclass) classification problems
\cite{beygelzimer05reductions}.  A cost-sensitive classification
example is given by an input $x$ and a cost vector $\vec c = \langle
c_1, \dots, c_K \rangle$, where $c_k$ is the cost of predicting class
$k$ on input $x$.  Define by \searn$(\cD^{\text{SP}}, \ell, \pi)$ a
distribution over cost-sensitive classification problems derived as
follows.  To sample from this induced distribution, we first sample an
example $(x,y) \sim \cD^{\text{SP}}$.  We then sample $t$ uniformly
from $[1,T]$ and run $\pi$ for $t-1$ steps on $(x,y)$.  This yields a
partial prediction $(\hat y_1, \dots, \hat y_{t-1})$.  The input for
the cost sensitive classification problem is then the tuple $(x,\hat
y_1, \dots, \hat y_{t-1})$.  The costs are derived as follows.  For
each possible choice $k$ of $\hat y_t$, we defined $c_k$ as the
\emph{expected} loss if $\pi$ were run, beginning at $(\hat y_1,
\dots, \hat y_{t-1}, k)$ on input $x$.  Formally:
\begin{equation}
c_k = \Ep_{\hat
  y_{t+1}, \dots, \hat y_T \sim \pi} \ell(y, (\hat y_1, \dots, \hat
y_{t-1}, k, \hat y_{t+1}, \dots, \hat y_T))
\end{equation}
\searn\ assumes access to an ``initial policy'' $\pi^*$ (sometimes
called the ``optimal policy'').  Given an input $x$, a true output $y$
and a prefix of predictions $\hat y_1, \dots, \hat y_{t-1}$, $\pi^*$
produces a best next-action, $\hat y_t$.  It should be constructed so
that the choice $\hat y_t$ is optimal (or close to optimal) with
respect to the problem-specific loss function.  For example, if the
loss function is Hamming loss, the $\pi^*$ will always produce $\hat
y_t = y_t$.  For more complex loss functions, computing $\pi^*$ may be
more involved.

\begin{figure}[t]
\framebox{
\begin{minipage}[t]{7.7cm}
~{\bf Algorithm} \searn-Learn$(\cA, \cD^{\text{SP}}, \ell, \pi^*, \be)$

\begin{algorithmic}[1]
\STATE Initialize $\pi = \pi^*$
\WHILE{not converged}
\STATE Sample: $D \sim $\searn$(\cD^{\text{SP}}, \ell, \pi)$
\STATE Learn:  $h \leftarrow \cA(D)$
\STATE Update: $\pi \leftarrow (1-\be) \pi + \be h$ 
\ENDWHILE
\STATE Return $\pi$ without reference to $\pi^*$
\end{algorithmic}
\end{minipage}
}
\caption{The complete \searn\ algorithm.  It's parameters are: a cost-sensitive classification algorithm $\cA$, a distribution over structured problems $\cD^{\text{SP}}$, a loss function $\ell$, an initial policy $\pi^*$ and an interpolation parameter $\be$.}
\label{fig:searn}
\end{figure}

Given these ingredients, \searn\ operates according the algorithm
given in Figure~\ref{fig:searn}.  Operationally, the sampling step is
typically implemented by generating \emph{every} example from a fixed
structured prediction training set.  The costs (expected losses) are
computed by sampling with tied randomness \cite{ng00pegasus}.

If $\be = 1/T^3$, one can show \cite{daume09searn} that after at most
$2T^3 \ln T$ iterations, \searn\ is guaranteed to find a
solution $\pi$ with structured prediction loss bounded as:
\begin{equation} \label{eq:searn}
L(\pi) \leq L(\pi^*) + 2 \ell_\textrm{avg} T \ln T + c (1+\ln T)/T
\end{equation}
where $L(\pi^*)$ is the loss of the initial policy (typically zero),
$T$ is the length of the longest example, $c$ is the worse-case
per-step loss and $\ell_\textrm{avg}$ is the average multiclass
classification loss.  This shows that the structured prediction
algorithm learned by \searn\ is guaranteed to be not-much-worse than
that produced by the initial policy, \emph{provided} that the created
classification problems are easy (i.e., that $\ell_\textrm{avg}$ is
small).  Note that one can use \emph{any} classification algorithm one
likes.

\section{Unsupervised \searn} \label{sec:unsearn}

In unsupervised structured prediction, we no longer receive an pair
$(x,y)$ but instead observes only an input $x$.  Our job is to
construct a classifier that produces $y$, even though we have never
observed it.

\subsection{Reduction for Unsupervised to Supervised}

The key idea---one that underlies much work in unsupervised
learning---is that a good $y$ is one that enables us to easily recover
$x$.  This is precisely the intuition we build in to our model.  The
observation that makes this practical is that there is nothing in the
theory or application of \searn\ that says that $\pi^*$ cannot be
stochastic.  Moreover, there is not requirement that the loss function
depend on \emph{all} components of the prediction.  Our model will
essentially first predict $y$ and then predict $x$ based on $y$.
Importantly, the loss function is agnostic to $y$ (since we do not
have true outputs).

The general construction is as follows. Let $\cD^{\text{unsup}}$ be a
distribution over inputs $x \in \cX$ and let $\cY$ be the space of
desired latent structures (e.g., trees). We define a distribution
$\cD^{\text{sup}}$ over $\cX \times (\cY \times \cX)$ by defining a
sampling procedure.  To sample from $\cD^{\text{sup}}$, we first
sample $x \sim \cD^{\text{unsup}}$.  We then sample uniformly from the
set of all $\cY$ that are valid structures for $x$.  Finally, we
return the pair $(x, (y, x))$.  We define a loss function $L$ by
$L((y, x), (\hat y, \hat x)) = L^{\text{input}}(x, \hat x)$ where
$L^{\text{input}}$ is any loss function on the input space (e.g.,
Hamming loss). We apply \searn\ to the supervised structured
prediction problem $\cD^{\text{sup}}$, and implicitly learn latent
structures.

\subsection{Sequence Labeling Example}

To gain insight into the operation of \searn\ in the unsupervised
setting, it is useful to consider a sequence labeling example.  That
is, our input $x$ is a sequence of length $T$ and we desire a label
sequence $y$ of length $T$ drawn from a label space of size $K$.  We
convert this into a supervised learning problem by considering the
``true'' structured output to be a label sequence of length $2T$, with
the first $T$ components drawn from the label space of size $K$ and
the second $T$ components drawn from the input vocabulary.  The loss
function can then be anything that depends only on the last $T$
components.  For simplicity, we can consider it to be Hamming loss.
The construction of the optimal policy in this case is
straightforward.  For the first $T$ components, $\pi^*$ may behave
arbitrarily (e.g., it may produce a uniform distribution over the $K$
labels).  For the second $T$ components, $\pi^*$ always predicts the
true label (which is known, because it is part of the input).

An important aspect of the model is the construction of the feature
vectors.  It is most useful to consider this construction as having
two parts.  The first part has to do with predicting the hidden
structure (the first $T$ components).  The second part has to do with
predicting the observed structure (the second $T$ components).  For
the first part, we are free to use whatever features we desire, so
long as they can be computed based on the input $x$ and a partial
output.  For instance, in the HMM case, we could use the two most
recent label predictions and windowed features from $x$.

The construction of the features for the second part is, however, also
crucial.  For instance, if the feature vector corresponding to
``predict the $t$th component of $x$'' contains the $t$ component of
$x$, then this learning problem is trivial---but also renders the
latent structure useless.  The goal of the designer of the feature
space is to construct features for predicting $x_t$ that crucially
depend on getting the latent structure $y$ correct.  That is, the
ideal feature set is one for which you can predict $x_t$ accurately
\emph{if an only if} we have found the correct latent structure (more
on this in Section~\ref{sec:guarantees}).  For instance, in the HMM
case, we may predict $x_t$ based only on the corresponding label
$y_t$, or maybe on the basis of $y_{t-1},y_t,y_{t+1}$.  (Note that we
are not limited to the Markov assumption, as in the case of HMMs.)

In the first iteration of \searn, all costs for the prediction of the
latent structure are computed with respect to the initial policy.
Recalling that the initial policy behaves randomly when predicting the
latent labels and correctly when predicting the words, we can see that
these costs are all \emph{zero}.  Thus, for the latent structure
actions, \searn\ will not induce any classification examples (because
the cost of all actions is equal).  However, it will create example
for predicting the $x$ component.  For predicting the $x$s, the cost
will be zero for the correct word and one for any incorrect word.
These examples will have associated features: we will predict word
$x_t$ based \emph{exclusively} on $y_t$.  Remember: $y_t$ was
generated randomly by the initial policy.

In the \emph{second} iteration, the behavior is different.  \searn\
returns to creating examples for the latent structure components.
However, in this iteration, since the current policy is not longer
optimal, the future cost estimates may be non-zero.  Consider
generating an example corresponding to a (latent) state $y_t$.  For
some small percentage (as dictated by $\beta$) of the ``generate $x$''
decisions, the previously learned classifier will fire.  If this
learned classifier does well, then the associated cost will be low.
However, if the learned classifier does poorly, the the associated
cost will be high.  Intuitively, the learned classifier will do well
if and only if the action that labels $y_t$ is ``good'' (i.e.,
consistent with what was learned previously).  This, in the second
pass through the data, \searn\ \emph{does} create classification
examples specific to the latent decisions.

As \searn\ iterates, more and more of the latent prediction decisions
are made according to the learned classifiers and not with respect to
the random policy.

\section{Comparison to EM} \label{sec:em}

In this section, we show an equivalence between expectation
maximization in directed probabilistic structures and unsupervised
\searn.  We use mixture of multinomials as a motivating example
(primarily for simplicity), but the results easily extend to more
complicated models (e.g., HMMs: see Section~\ref{sec:hmm}).

\subsection{EM for Mixture of Multinomials} \label{sec:em-mm}

In the mixture of multinomials problem, we are given $N$ documents
$\vec d_1, \dots, \vec d_N$, where $\vec d_n$ is a vector of word
counts over a vocabulary of size $V$; that is, $d_{n,v}$ is the number
of times word $v$ appeared in document $n$.  The mixture of
multinomials is a probabilistic clustering model, where we assume an
underlying set of $K$ clusters (multinomials) that generated the
documents.  Denote by $\th_k$ the multinomial parameter associated
with cluster $k$, $\rho_k$ the prior probability of choosing cluster
$k$, and let $\vec z_n$ be an indicator vector associating document
$n$ with the unique cluster $k$ such that $z_{n,k} = 1$.  The
probabilistic model has the form:
\begin{equation} \label{eq:mmlik}
  p(\vec d \| \vec \th, \vec \rho) =
    \prod_n 
      \frac {(\sum_v d_{n,v})!}
            {\prod_v d_{n,v}!}
        \sum_{\vec z_n} \prod_k \left[ 
          \rho_k
          \prod_v \th_{k,v}^{d_{n,v}}
          \right]^{z_{n,k}}
\end{equation}
Expectation maximization in this model involves first computing
expectations over the $\vec z$ vectors and then updating the model
parameters $\vec \th$:
\begin{small}
\begin{align}
\text{E-step:} &&
z_{n,k} &\varpropto \rho_k \prod_v \th_{k,v}^{d_{n,v}} \label{eq:em:z} \\
\text{M-step:} &&
\th_{k,v} &\varpropto \sum_n z_{n,k} d_{n,v} 
&
;~~~ \rho_k &\varpropto \sum_n z_{n,k} \label{eq:em:rho}
\end{align}
\end{small}
In both cases, the constant of proportionality is chosen so that the
variables sum to one over the last component.  These updates are
repeated until convergence of the incomplete data likelihood,
Eq~\eqref{eq:mmlik}.

\subsection{An Equivalent Model in Searn} \label{sec:em-searn}

Now, we show how to construct an instance of unsupervised \searn\ that
effectively mimics the behavior of EM on the mixture of multinomials
problem.  The ingredients are as follows:
\begin{small}
\begin{itemize}
\item The input space $\cX$ is the space of documents, represented as word count vectors.
\item The (latent) output space $\cY$ is a single discrete variable in the range $[1,K]$ that specifies the cluster.
\item The feature set for predicting $y$ (document counts).
\item The feature set for predicting $x$ is the label $y$ and the total number of words in the document.  The predictions for a document are estimated word probabilities, not the words themselves.
\item The loss function ignores the prediction $y$ and returns the log loss of the true document $x$ under the word probabilities predicted.
\item The cost-sensitive learning algorithm is different depending on whether the latent structure $y$ is being predicted or if the document $x$ is being predicted:
\begin{itemize}
\item Structure: The base classifier is a multinomial na\"ive Bayes
  classifier, parameterized by (say) $h^m$
\item Document:  The base classifier is a collection of independent
  maximum likelihood multinomial estimators for each cluster.
\end{itemize}
\end{itemize}
\end{small}
Consider the behavior of this setup.  In particular, consider the
distribution \searn$(\cD^{\text{SP}}, \ell, \pi)$.  There are two
``types'' of examples drawn from this distribution: (1) latent
structure examples and (2) document examples.  The claim is that
\emph{both} classifiers learned are identical to the mixture of
multinomials model from Section~\ref{sec:em-mm}.

Consider the generation of a latent structure example.  First, a
document $n$ is sampled uniformly from the training set.  Then, for
each possible label $k$ of this document, a cost $\Ep_{\vec {\hat d}
  \sim \pi} l((y,\vec d_n), (k,\vec {\hat d}))$ is computed.  By
definition, the $\vec {\hat d}$ that is computed is exactly the
prediction according to the current multinomial estimator, $h^m$.
Interpreting the multinomial estimator in terms of the EM parameters,
the costs are \emph{precisely} the $z_{n,k}$s from EM (see
Eq~\eqref{eq:em:z}).  These latent structure examples are fed in to
the multinomial na\"ive Bayes classifier, which re-estimates a model
exactly as per the M-step in EM (Eq~\eqref{eq:em:rho}).

Next, consider the generation of the document examples.  These
examples are generated by $\pi$ first choosing a cluster according to
the structure classifier.  This cluster id is then used as the (only)
feature to the ``generate document'' multinomial.  As we saw before,
the probability that $\pi$ will select label $k$ for document $n$ is
precisely $z_{n,k}$ from Eq~\eqref{eq:em:z}.  Thus, the multinomial
estimator will effectively receive weighted examples, weighted by
these $z_{n,k}$s, thus making the maximum likelihood estimate exactly
the same as the M-step from EM (Eq~\eqref{eq:em:rho}).

\subsection{Synthetic experiments} \label{sec:hmm}

To demonstrate the advantages of the generality of \searn, we report
here the result of some experiments on synthetic data.  We generate
synthetic data according to two different HMMs.  The first HMM is a
first-order model.  The initial state probabilities, the transition
probabilities, and the observation probabilities are all drawn
uniformly.  The second HMM is a second-order model, also will all
probabilities drawn uniformly. The lengths of observations are given
by a Poisson with a fixed mean.

\begin{table*}[t]
\centering
\caption{Error rates on first- and second-order Markov data with 2, 5 or 10 latent states. Models
are the true data generating distribution (approximated by a first-order Markov model in the case of
âHMM2â), a model learned by EM, one learned by \searn\  with a na\"ive Bayes base classifier, and
one learned by \searn\  with a logistic regression base classifier. Standard deviations are given in
small text. The best results by row are bolded; the results within the standard deviation of the best
results are italicized.}
\begin{tabular}{|l|l|c|c|cc|}
\hline
Model &States &Truth &EM &\searn\ -NB &\searn\ -LR \\
\hline
1st order HMM  &K = 2  &0.227 {\tiny $\pm 0.107$} &\it 0.275 {\tiny $\pm 0.128$} &\bf 0.287 {\tiny $\pm 0.138$} &\it 0.276 {\tiny $\pm 0.095$} \\
1st order HMM  &K = 5  &0.687 {\tiny $\pm 0.043$} &\it 0.678 {\tiny $\pm 0.026$} &\it 0.688 {\tiny $\pm 0.025$} &\bf 0.672 {\tiny $\pm 0.022$} \\
1st order HMM  &K = 10 &0.806 {\tiny $\pm 0.035$} &\it 0.762 {\tiny $\pm 0.021$} &\it 0.771 {\tiny $\pm 0.019$} &\bf 0.755 {\tiny $\pm 0.019$} \\
\hline
2nd order HMM  &K = 2  &0.294 {\tiny $\pm 0.072$} &0.396 {\tiny $\pm 0.057$} &0.408 {\tiny $\pm 0.056$} &\bf 0.271 {\tiny $\pm 0.057$} \\
2nd order HMM  &K = 5  &0.651 {\tiny $\pm 0.068$} &0.695 {\tiny $\pm 0.027$} &0.710 {\tiny $\pm 0.016$} &\bf 0.633 {\tiny $\pm 0.018$} \\
2nd order HMM  &K = 10 &0.815 {\tiny $\pm 0.032$} &0.764 {\tiny $\pm 0.021$} &0.771 {\tiny $\pm 0.015$} &\bf 0.705 {\tiny $\pm 0.019$} \\
\hline
\end{tabular}
\label{tab:hmm}
\end{table*}

In our experiments, we consider the following learning algorithms: EM,
\searn\ with HMM features and a na\"ive Bayes classifier, and \searn\
with a logistic regression classifier (and an enhanced feature space:
predicting $y_t$ depends on $x_{t-1:t+1}$. The first \searn\ should
mimic EM, but by using sampling rather than exact expectation
computations.  The models are all first-order, regardless of the
underlying process.

We run the following experiment.  For a given number of states (which
we will vary), we generate $10$ random data sets according to each
model.  Each data set consists of $5$ examples with mean example
length of $40$ observations. The vocabulary size of the observed data
is always $10$. We compute error rates by matching each predicted
label to the best-matching true label and the compute Hamming loss.
Forward-backward is initialized randomly.  We run experiments with the
number of latent states equal to $2$, $5$ and $10$.\footnote{We ran
  experiments varying the number of samples \searn\ uses in $\{1, 2,
  5\}$; there was no statistically significant difference. The
  results we report are based on $2$ samples.}

The results of the experiments are shown in Table~\ref{tab:hmm}.  The
observations show two things.  When the true model matches the model
we attempt to learn (HMM1), there is essentially no statistically
significant difference between any of the algorithms. Where once sees
a difference is when the true model does not match the learned model
(HMM2). In this case, we see that \searn-LR obtains a significant
advantage over both EM and \searn-NB, due to its ability to employ a
richer set of features. These results hold over all values of $K$.
This is encouraging, since in the real world our model is rarely (if
ever) right.  The (not statistically significant) difference in error
rates between EM and \searn-NB are due to a sampling versus exact
computation of expectations.  Many of the models outperform ``truth''
because likelihood and accuracy do not necessarily correlate
\cite{liang08errors}.

\section{Analysis} \label{sec:guarantees}

There are two keys to success in unsupervised-\searn. The first key is
that the features on the $\cY$-component of the output space be
descriptive enough that it be learnable.  One way of thinking of this
constraint is that if we had labeled data, then we would be able to
learn well.  The second key is that the features on the
$\cX$-component of the output space be intrinsically tied to the hidden
component.  Ideally, these features will be such that $\cX$ can be
predicted with high accuracy if and only if $\cY$ is predicted
accurately.

The general--though very trivial--result is that if we can guarantee
that the loss on $\cY$ is bounded by some function $f$ of the loss on
$\cX$, then the loss on $\cY$ is guaranteed after learning to be
bounded by $f(L(\pi^*) + 2\ell_{\text{avg}}T_{\text{max}} ln
T_{\text{max}} + c(1 + ln T_{\text{max}})/T_{\text{max}})$, where all
the constants now depend on the induced structured prediction problem;
see Eq~\ref{eq:searn}.

One can see the unsupervised \searn\ analysis as justifying a small
variant on ``Viterbi training''--the process of performing EM where
the E-step is approximated with a delta function centered at the
maximum.  One significant issue with Viterbi training is that it is
not guaranteed to converge.  However, Viterbi training is recovered as
a special case of unsupervised \searn\ where the interpolation
parameter is fixed at $1$.  While the \searn\ theorem no longer
applies in this degenerate case, any algorithm that uses Viterbi
training could easily be retrofitted to simply make some decisions
randomly.  In doing so, one would obtain an algorithm that does have
theoretical guarantees.

\section{Unsupervised Dependency Parsing} \label{sec:gi}

\begin{figure}
\centering
\includegraphics[width=0.45\textwidth]{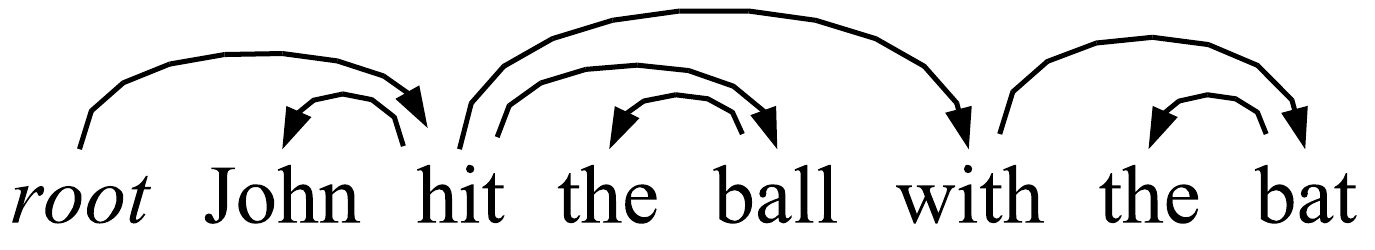}
\caption{Dependency parse of a $T=7$ word sentence.}
\label{fig:deptree}
\end{figure}

The dependency formalism is a practical and linguistically interesting
model of syntactic structure. One can think of a dependency structure
for a sentence of length $T$ as a directed tree over a graph over $T +
1$ nodes: one node for each word plus a unique root node. Edges point
from heads to dependents. An example dependency structure for a $T =
7$ word sentence is shown in Figure~\ref{fig:deptree} . To date,
unsupervised dependency parsing has only been viewed in the context of
global probabilistic models specified over dependency pairs
\cite{paskin01bigrams} or spanning trees
\cite{klein04induction,smith05induction}.  However, there is an
alternative, popular method for producing dependency trees in a
supervised setting: shift-reduce parsing
\cite{nivre03parsing,sagae05shiftreduce}.

\subsection{Shift-reduce dependency parsing}

Shift-reduce dependency parsing \cite{nivre03parsing} is a
left-to-right parsing algorithm that operates by maintaining three
state variables: a stack $S$, a current position $i$ and a set of arcs
$A$. The algorithm begins with $\langle S, i,A\rangle = \langle
\emptyset, 1, \emptyset \rangle$: the stack and arcset are empty and
the current index is $1$ (the first word).  The algorithm then
proceeds through a series of actions until a final state is reached. A
final state is one in which $i = T$, at which point the set $A$
contains all dependency edges for the parse.  Denote by $i|I$ a stack
with $i$ at the head and stack $I$ at the tail.  There are four
actions:

\vspace{-2mm}
\begin{small}
\begin{description}
\item[LeftArc:] $\langle t|S, i, A \rangle \longrightarrow \langle S,
  i, (i, t)|A \rangle$, so long as there does not exist an arc
  $(\cdot·, t) \in A$.  (Adds a left dependency to the arc set between
  the word $t$ at the top of the stack and the word $i$ at the current
  index.)

\item[RightArc:] $\langle t|S, i, A \rangle \longrightarrow \langle
  i|t|s, i + 1, (t, i)|A\rangle$, so long as there is no arc $(\cdot·,
  i) \in A$.  (Adds a right dependency between the top of the stack
  and the next input.)

\item[Reduce:] $\langle t|S, i,A \rangle \longrightarrow \langle S,
  i,A \rangle$, so long as there does exist an arc $(\cdot·, t) \in
  A$. (Removes a word from the stack.)

\item[Shift:] $\langle S, i, A\rangle \longrightarrow \langle n|S, i +
  1,A \rangle$. (Place item on stack.)
\end{description}
\end{small}
\vspace{-2mm}

This algorithm is guaranteed to terminate in at most $2T$ steps with a
valid dependency tree \cite{nivre03parsing}, unlike standard
probabilistic algorithms that have a time-complexity that is cubic in
$T$ \cite{mcdonald07complexity}.  The advantage of the shift-reduce
framework is that it fits nicely into \searn. However, until now, it
has been an open question how to train a shift-reduce model in an
unsupervised fashion. The techniques described in this paper give a
solution to this problem.

\subsection{Experimental setup}

We follow the same experimental setup as \cite{smith05induction},
using data from the WSJ10 corpus (sentences of length at most ten from
the Penn Treebank \cite{marcus93treebank}). The data is stripped of
punctuation and parsing depends on the part-of-speech tags, not the
words. We use the same train/dev/test split as Smith and Eisner:
$5301$ sentences of training data, $531$ sentences of development data
and $530$ sentences of blind test data.  All algorithm development and
tuning was done on the development data.

We use a slight modification to SearnShell to facilitate the
development of our algorithm together with a multilabel logistic
regression classifier, MegaM.\footnote{SearnShell and MegaM are
  available at \url{http://searn.hal3.name} and
  \url{http://hal3.name/megam}, respectively.} Our algorithm uses the
following features for the tree-based decisions (inspired by
\cite{hall06dependency}), where $t$ is the top of the stack and $i$ is
the next token: the parts-of-speech within a window of $2$ around $t$
and $i$; the pair of tokens at $t$ and $i$; the distance (discretized)
between $t$ and $i$; and the part-of-speech at the head (resp. tail)
of any existing arc pointing to (resp. from) $t$ or $i$. For producing
word $i$, we use the part of speech of $i$'s parent, grandparent,
daughters and aunts.

We use \searn\ with a fixed $\be = 0.1$.  One sample is used to
approximate expected losses.  The development set is used to tune the
scale of the prior variances for the logistic regression (different
variances are allowed for the ``produce tree'' and ``produce words''
features). The initial policy makes uniformly random decisions.
Accuracy is directed arc accuracy.

\subsection{Experimental results}

\begin{table}[t]
\centering
\caption{Accuracy on training and test data, plus number of iterations for a variety of dependency
parsing algorithms (all unsupervised except for the last two rows).}
\begin{tabular}{|l|c|c|r|}
\hline
Algorithm & Acc-Tr & Acc-Tst & \# Iter \\
\hline
Rand-Gen       & 23.5 {\tiny $\pm 0.9$}& 23.5 {\tiny $\pm 1.3$ }& \\
Rand-\searn\   & 21.3 {\tiny $\pm 0.2$}& 21.0 {\tiny $\pm 0.6$} & \\
\hline
K+M:Rand-Init  & 23.6 {\tiny $\pm 3.8$}& 23.6 {\tiny $\pm 4.3$}& 63.3 \\
K+M:Smart-Init & 35.2 {\tiny $\pm 6.6$}& 35.2 {\tiny $\pm 6.0$}& 64.1 \\
\hline
S+E:Length     & 33.8 {\tiny $\pm 3.6$}& 33.7 {\tiny $\pm 5.9$}& 173.1 \\
S+E:DelOrTrans1& 47.3 {\tiny $\pm 6.0$}& 47.1 {\tiny $\pm 5.9$}& 132.2 \\
S+E:Trans1     & 48.8 {\tiny $\pm 0.9$}& 49.0 {\tiny $\pm 1.5$}& 173.4 \\
\hline
\searn: Unsup  & 45.8 {\tiny $\pm 1.6$}& 45.4 {\tiny $\pm 2.2$}& 27.6 \\
\hline
S+E: Sup       & 79.9 {\tiny $\pm 0.2$}& 78.6 {\tiny $\pm 0.8$}& 350.5 \\
\searn: Sup    & 81.0 {\tiny $\pm 0.3$}& 81.6 {\tiny $\pm 0.4$}& 24.4 \\
\hline
\end{tabular}
\label{tab:dep}
\end{table}

The baseline systems are: two random baselines (one generative, one
given by the \searn\ initial policy), Klein and Manning's model
\cite{klein04induction} EM-based model (with and without clever
initialization), and three variants of Smith and Eisner's model
\cite{smith05induction} (with random initialization, which seems to be
better for most of their models). We also report an ``upper bound''
performance based on supervised training, for both the probabilistic
(Smith+Eisner model) as well as supervised \searn.

The results are reported in Table~\ref{tab:dep}: accuracy on the
training data, accuracy on the test data and the number of iterations
required. These are all averaged over $10$ runs; standard deviations
are shown in small print. Many of the results (the non-\searn\
results) are copied from \cite{smith05induction}. The stopping criteria for the
EM-based models is that the log likelihood changes by less than
$10e-5$.  For the \searn-based methods, the stopping criteria is that
the development accuracy ceases to increase (on the individual
classification tasks, not on the structured prediction task).

All learned algorithms outperform the random algorithms (except
Klein+Manning with random inits).  K+M with smart initialization does
slightly better than the worst of the S+E models, though the
difference is not statistically significant.  It does so needing only
about a third of the number of iterations (moreover, a single
S+E iteration is slower than a single K+M
iteration). The other two S+E models do roughly comparably in
terms of performance (strictly dominating the previous methods). One
of them (``DelOrTrans1'') requires about twice as many iterations as
K+M; the other (``Trans1'') requires about three times (but
has much high performance variance). Unsupervised \searn\ performs
halfway between the best K+M model and the best S+E model (it is
within the error bars for ``DelOrTrans1'' but not ``Trans1'').

Nicely, it takes significantly fewer iterations to converge (roughly
$15\%$). Moreover, each iteration is quite fast in comparison to the
EM-based methods (a complete run took roughly $3$ hours on a 3.8GHz
Opteron using SearnShell). Finally, we present results for the
supervised case. Here, we see that the \searn-based method converges
much more quickly to a better solution than the S+E model.
Note that this comparison is unfair since the \searn-based model uses
additional features (though it is a nice property of the \searn-based
model that it \emph{can} make use of additional features).
Nevertheless we provide it so as to give a sense of a reasonable
upper-bound. We imagine that including more features would shift the
upper-bound and the unsupervised algorithm performance up.

\section{A Semi-Supervised Version} \label{sec:semi}

\begin{figure}
\centering
\includegraphics[width=0.4\textwidth]{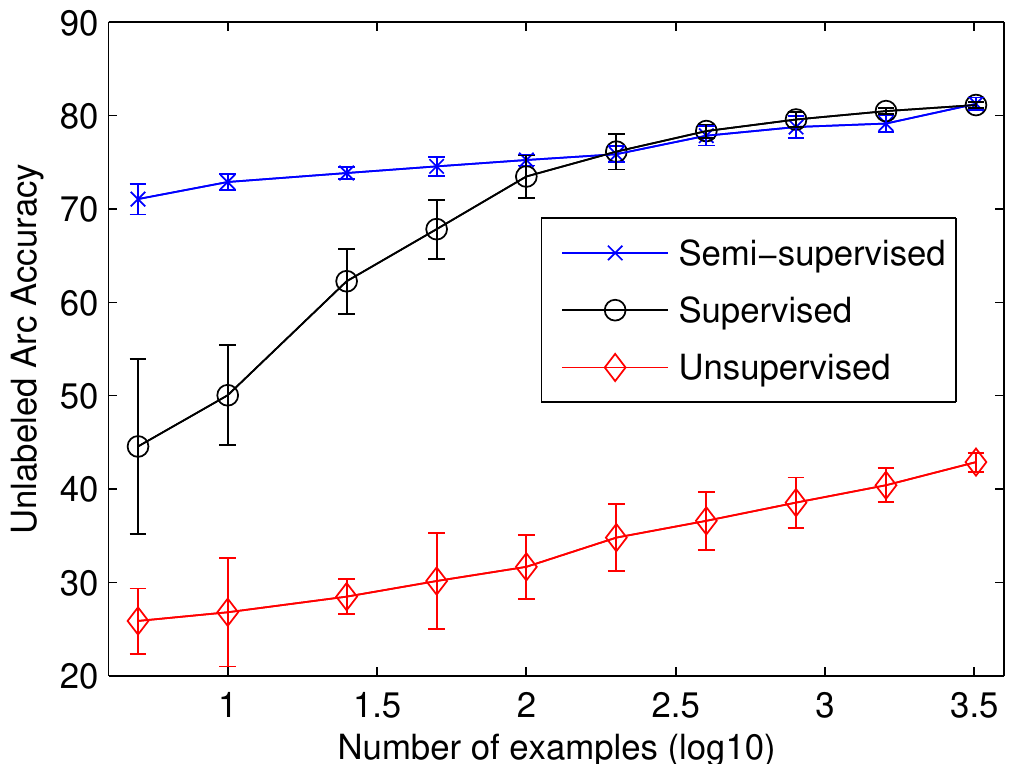}
\caption{Parsing accuracy for semi-supervised, supervised and unsupervised \searn.  X-axis is: (semi/sup) \# of labeled examples; (unsup) \# of unlabeled examples.}
\label{fig:semisup}
\end{figure}

The unsupervised learning algorithm described above naturally extends
to the case where some labeled data is available.  In fact, the only
modification to the algorithm is to change the loss function.  In the
unsupervised case, the loss function completely ignores the latent
structure, and returns a loss dependent only on the ``predict self''
task.  In the semi-supervised version, one plugs in a natural loss
function for the ``latent'' structure prediction for the labeled
subset of the data.

In Figure~\ref{fig:semisup}, we present results on dependency parsing.
We show learning curves for unsupervised, fully supervised and
semi-supervised models.  The x-axis shows the number of examples used;
in the unsupervised and supervised cases, this is the total number of
examples; in the semi-supervised case, it is the number of labeled
examples.  Error bars are two standard deviations.  Somewhat
surprisingly, with only five labeled examples, the semi-supervised
approach achieves an accuracy of over $70\%$, only about $10\%$ behind
the fully supervised approach with $5182$ labeled examples.
Eventually the supervised model catches up (at about $250$ labeled
examples).  The performance of the unsupervised model continues to
grow as more examples are provided, but never reaches anywhere close
to the supervised or semi-supervised models.

\section{Conclusions}

We have described the application of a search-based structured
prediction algorithm, \searn, to unsupervised learning. This answers
positively an open question in the field of learning reductions
\cite{beygelzimer05reductions}: can unsupervised learning be reduced
to supervised learning?  We have shown a near-equivalence between the
resulting algorithm and the forward-backward algorithm in hidden
Markov models. We have shown an application of this algorithm to
unsupervised dependency parsing in a shift-reduce framework. This
provides the first example of unsupervised learning for dependency
parsing in a non-probabilistic model and shows that unsupervised
shift-reduce parsing is possible.  One obvious extension of this work
is to structured prediction problems with additional latent structure,
such as in machine translation.  Instead of using the predict-self
methodology, one could directly apply a predict-target methodology.

The view of ``predict the input'' for unsupervised learning is
implicit in many unsupervised learning approaches, including standard
models such as restricted Boltzmann machines and Markov random fields.
This is made most precise in the wake-sleep algorithm
\cite{hinton95wakesleep}, which explicitly trains a neural network to
reproduce its own input.  The wake-sleep algorithm consists of two
phases: the wake phase, where the latent layers are produced, and the
sleep phase, where the input is (re-)produced.  These two phases are
analogous to the predict-structure phase and the predict-words phase
in unsupervised \searn.

\vspace{-1mm}
\paragraph{Acknowledgements.}
Thanks for Ryan McDonald and Joakim Nivre for discussions related to
dependency parsing algorithms.  Comments from 5 (!)  anonymous
reviewers were incredibly helpful.  This was partially supported by
NSF grant IIS-0712764.


\vspace{-1.4mm}
\begin{thebibliography}{}

\vspace{-0.7mm}
\bibitem[Beygelzimer et~al.\/, 2005][Beygelzimer
  et~al.\/][2005]{beygelzimer05reductions}
Beygelzimer, A., Dani, V., Hayes, T., Langford, J., \& Zadrozny, B. (2005).
\newblock Error limiting reductions between classification tasks.
\newblock {\em Proc. Int'l Conf. on Machine Learning
  } (pp.\/ 49--56).

\vspace{-0.7mm}
\bibitem[Brown et~al.\/, 1993][Brown et~al.\/][1993]{brownetal93}
Brown, P., {Della Pietra}, S., {Della Pietra}, V., \& Mercer, R. (1993).
\newblock The mathematics of statistical machine translation: Parameter
  estimation.
\newblock {\em Computational Linguistics}, {\em 19}, 263--311.

\vspace{-0.7mm}
\bibitem[{Daum\'e III} et~al.\/, 2009 to appear][{Daum\'e III} et~al.\/][2009
  (to appear)]{daume09searn}
{Daum\'e III}, H., Langford, J., \& Marcu, D. (2009 (to appear)).
\newblock Search-based structured prediction.
\newblock {\em Machine Learning J.}.

\vspace{-0.7mm}
\bibitem[Dempster et~al.\/, 1977][Dempster et~al.\/][1977]{dempster77em}
Dempster, A., Laird, N., \& Rubin, D. (1977).
\newblock Maximum likelihood from incomplete data via the {EM} algorithm.
\newblock {\em J. of the Royal Statistical Society}, {\em B39}, 1--38.

\vspace{-0.7mm}
\bibitem[Hall et~al.\/, 2006][Hall et~al.\/][2006]{hall06dependency}
Hall, J., Nivre, J., \& Nilsson, J. (2006).
\newblock Discriminative classifiers for determining dependency parsing.
\newblock {\em Proc. Conf. of the Assoc. for
  Computational Linguistics} (pp.\/ 316--323).

\vspace{-0.7mm}
\bibitem[Hinton et~al.\/, 1995][Hinton et~al.\/][1995]{hinton95wakesleep}
Hinton, G., Dayan, P., Frey, B., \& Neal, R. (1995).
\newblock The wake-sleep algorithm for unsupervised neural networks.
\newblock {\em Science}, {\em 26}, 1158--1161.

\vspace{-0.7mm}
\bibitem[Klein \& Manning, 2004][Klein and Manning][2004]{klein04induction}
Klein, D., \& Manning, C. (2004).
\newblock Corpus-based induction of syntactic structure: Models of dependency
  and constituency.
\newblock {\em Proc. Conf. of the Assoc. for
  Computational Linguistics} (pp.\/ 478--485).

\vspace{-0.7mm}
\bibitem[Lafferty et~al.\/, 2001][Lafferty et~al.\/][2001]{lafferty01crf}
Lafferty, J., McCallum, A., \& Pereira, F. (2001).
\newblock Conditional random fields: Probabilistic models for segmenting and
  labeling sequence data.
\newblock {\em Proc. Int'l Conf. on Machine Learning
  } (pp.\/ 282--289).

\vspace{-0.7mm}
\bibitem[Liang \& Klein, 2008][Liang and Klein][2008]{liang08errors}
Liang, P., \& Klein, D. (2008).
\newblock Analyzing the errors of unsupervised learning.
\newblock {\em Proc. Assoc. for Computational
  Linguistics} (pp.\/ 879--887).

\vspace{-0.7mm}
\bibitem[Marcus et~al.\/, 1993][Marcus et~al.\/][1993]{marcus93treebank}
Marcus, M., Marcinkiewicz, M.~A., \& Santorini, B. (1993).
\newblock Building a large annotated corpus of {E}nglish: The {P}enn
  {T}reebank.
\newblock {\em Computational Linguistics}, {\em 19}, 313--330.

\vspace{-0.7mm}
\bibitem[McDonald \& Satta, 2007][McDonald and
  Satta][2007]{mcdonald07complexity}
McDonald, R., \& Satta, G. (2007).
\newblock On the complexity of non-projective data-driven dependency parsing.
\newblock {\em Int'l Wk. on Parsing Technologies} (pp.\/
  121--132).

\vspace{-0.7mm}
\bibitem[Ng \& Jordan, 2000][Ng and Jordan][2000]{ng00pegasus}
Ng, A., \& Jordan, M. (2000).
\newblock {PEGASUS}: A policy search method for large {MDP}s and {POMDP}s.
\newblock {\em Proc. Converence on Uncertainty in Artificial
  Intelligence} (pp.\/ 406--415).

\vspace{-0.7mm}
\bibitem[Nivre, 2003][Nivre][2003]{nivre03parsing}
Nivre, J. (2003).
\newblock An efficient algorithm for projective dependency parsing.
\newblock {\em Int'l Wk. on Parsing Technologies} (pp.\/
  149--160).

\vspace{-0.7mm}
\bibitem[Paskin, 2001][Paskin][2001]{paskin01bigrams}
Paskin, M.~A. (2001).
\newblock Grammatical bigrams.
\newblock {\em Advances in Neural Info. Processing Systems} (pp.\/
  91--97).

\vspace{-0.7mm}
\bibitem[Rabiner, 1989][Rabiner][1989]{rabiner89tutorial}
Rabiner, L. (1989).
\newblock A tutorial on hidden {Markov} models and selected applications in
  speech recognition.
\newblock {\em Proc. IEEE} (pp.\/ 257--285).

\vspace{-0.7mm}
\bibitem[Ramanan et~al.\/, 2005][Ramanan et~al.\/][2005]{ramanan05pose}
Ramanan, D., Forsyth, D., \& Zisserman, A. (2005).
\newblock Strike a pose: Tracking people by finding stylized poses.
\newblock {\em Computer Vision and Pattern Recognition} (pp.\/
  271--278).

\vspace{-0.7mm}
\bibitem[Sagae \& Lavie, 2005][Sagae and Lavie][2005]{sagae05shiftreduce}
Sagae, K., \& Lavie, A. (2005).
\newblock A classifier-based parser with linear run-time complexity.
\newblock {\em Int'l Wk. on Parsing Technologies}.

\vspace{-0.7mm}
\bibitem[Smith \& Eisner, 2005][Smith and Eisner][2005]{smith05induction}
Smith, N.~A., \& Eisner, J. (2005).
\newblock Guiding unsupervised grammar induction using contrastive estimation.
\newblock {\em IJCAI Wk. on Grammatical Inference Apps} (pp.\/
  73--82).

\vspace{-0.7mm}
\bibitem[Taskar et~al.\/, 2005][Taskar et~al.\/][2005]{taskar05mmmn}
Taskar, B., Chatalbashev, V., Koller, D., \& Guestrin, C. (2005).
\newblock Learning structured prediction models: A large margin approach.
\newblock {\em Proc. Int'l Conf. on Machine Learning
  } (pp.\/ 897--904).

\vspace{-0.7mm}
\bibitem[Titov \& Henderson, 2007][Titov and Henderson][2007]{titov07latent}
Titov, I., \& Henderson, J. (2007).
\newblock A latent variable model for generative dependency parsing.
\newblock {\em Int'l Conf. on Parsing Technologies}.

\vspace{-0.7mm}
\bibitem[Vogel et~al.\/, 1996][Vogel et~al.\/][1996]{vogel96}
Vogel, S., Ney, H., \& Tillmann, C. (1996).
\newblock {HMM}-based word alignment in statistical translation.
\newblock {\em Proc. Int'l Conf. on Computational
  Linguistics} (pp.\/ 836--841).

\end{thebibliography}

\end{document}